\documentclass[preprint,review,12pt,number]{elsarticle}
\usepackage{url}
\usepackage{amssymb}
\usepackage{amsmath}
\usepackage{caption}
\usepackage{booktabs}
\usepackage{graphicx}

\journal{Nuclear Physics B}

\begin{document}

\begin{frontmatter}



\title{TCLA: Training-Free Class-wise Logit Adaptation for Medical Vision-Language Models} %% Article title


\author[2]{Tianyou Jiang} \ead{tianyou.jiang.research@gmail.com}
\author[3]{Ziyu Zhou}
% \author[1,2]{Haozhe Luo\corref{cor1}}
% \ead{haozhe.luo@unibe.ch}
% \author[1,2]{Mauricio Reyes\corref{cor1}}
% \ead{mauricio.reyes@unibe.ch}

% \cortext[cor1]{Corresponding author.}
% \cortext[cor1]{Corresponding author.}

% \affiliation[1]{
%   organization={ARTORG Center for Biomedical Engineering Research},
%   city={Bern},
%   country={Switzerland}
% }
\affiliation[2]{
  organization={University of Bern},
  city={Bern},
  country={Switzerland}
}
\affiliation[3]{
  organization={Shanghai Jiao Tong University},
  city={Shanghai},
  country={China}
}
\vspace{-1.5em}
%% Abstract
\begin{abstract}
Medical Vision–Language Models (VLMs) exhibit strong zero-shot performance, yet their effectiveness still declines on out-of-distribution (OOD) data due to domain shifts and class bias inherited from large-scale pretraining. Existing few-shot adaptation methods typically introduce additional trainable components, which can be unstable in extremely low-data regimes (e.g., 1-shot), and lack robustness on different medical data. We present TCLA, a purely training-free few-shot adaptation method for Medical VLMs, which is fast and model-agnostic. TCLA corrects inference logits based on a small set of support samples, boosting pretrained VLMs performance by improving inter-class deconfusion and reducing domain shift. Extensive experiments on nine datasets across multiple medical imaging modalities including X-ray, Ultrasound, MRI, CT, Histopathology, demonstrate that TCLA consistently improves OOD performance of Medical VLMs and, in most of cases, outperforms existing training-based adaptation methods.
\end{abstract}

% \centering
% \vspace{0.5em}
% \includegraphics[width=1.06\textwidth]{GraphicAbstract.pdf}

% \begin{highlights}
% \item TCLA adapts pretrained Medical Vision-Language Models using few-shot support samples without training.
% \item TCLA effectively and interpretably improves zero-shot performance on downstream diagnostic tasks by correcting zero-shot logits, thus mitigating the effects of domain shift in medical images.
% \item TCLA achieves robust improvements across nine medical datasets compared with both training-free and training-based methods.
% \end{highlights}

%% Keywords

\end{frontmatter}



\section{Introduction}
Vision and Language Models (VLMs) have attracted increasing attention following the success of CLIP (Contrastive Language Image Pretraining), which maps images and text into a shared embedding space \cite{Radford2021}. The growing availability of large-scale image–caption datasets in the medical domain has further enabled Vision–Language Pretraining (VLP) for medical applications. Several Medical CLIP variants, including PubMedCLIP \cite{Eslami2023}, MedCLIP \cite{Wu2023}, BioMedCLIP \cite{Zhang2023BioMedCLIP}, and BIOMEDICA \cite{Lozano2025}, have demonstrated strong zero-shot (ZS) performance on downstream medical imaging tasks, such as disease diagnosis \cite{Luo2024}, medical image segmentation \cite{Aleem2024}, and radiology report generation \cite{Liu2024}. 

Nevertheless, real-world medical data are often out-of-distribution (OOD), i.e., unseen disease categories present different image distributions due to variations in hospitals, scanners, or patient populations. Medical CLIP-based models can exhibit poor Zero-Shot performance when processing OOD datasets due to domain shifts and inherent class imbalance in their large-scale pretraining dataset \cite{Zhong2025}\cite{Koleilat2025BiomedCoOp}. A recent comparative study between general-purpose CLIP and Medical CLIP reported that Medical CLIP models may offer only limited advantages under transfer learning settings on OOD benchmarks \cite{Zhong2025}. 

Various few-shot learning (FS) methods have been proposed to adapt general VLMs for improved performance. These approaches typically freeze the image and text encoders, introduce lightweight adaptive modules, and employ strategies such as Prompt Learning (PL) \cite{Zhou2022,Yao2023}, Feature Adaptation (FA) \cite{Hu2022,Gao2024}, or Logit Adaptation (LA) \cite{Li2025,Zhang2022,Bendou2025Proker}. However, rapid adaptation with extremely limited labeled data has been reported to be unstable under out-of-distribution (OOD) conditions. A recent comparative analyze of few-shot adaptation methods for medical VLMs shows that conventional Linear Probing (LP) \cite{Yosinski2014} can outperform more complex adaptation strategies in certain low-shot regimes \cite{Kravets2025}. Suggesting that adaptive parameters become difficult to reliably optimize and adapt Medical VLMs when only a handful of training samples (e.g., 1-shot) are available. In such scenarios, training-free approaches may offer advantages in stability, computational efficiency, and deployment simplicity.

\begin{figure}[t]
\centering 
\includegraphics[width=1\textwidth, 
                 height=1\textheight, 
                 keepaspectratio,    
                 trim=0cm 20cm 0cm 0cm,
                 clip]{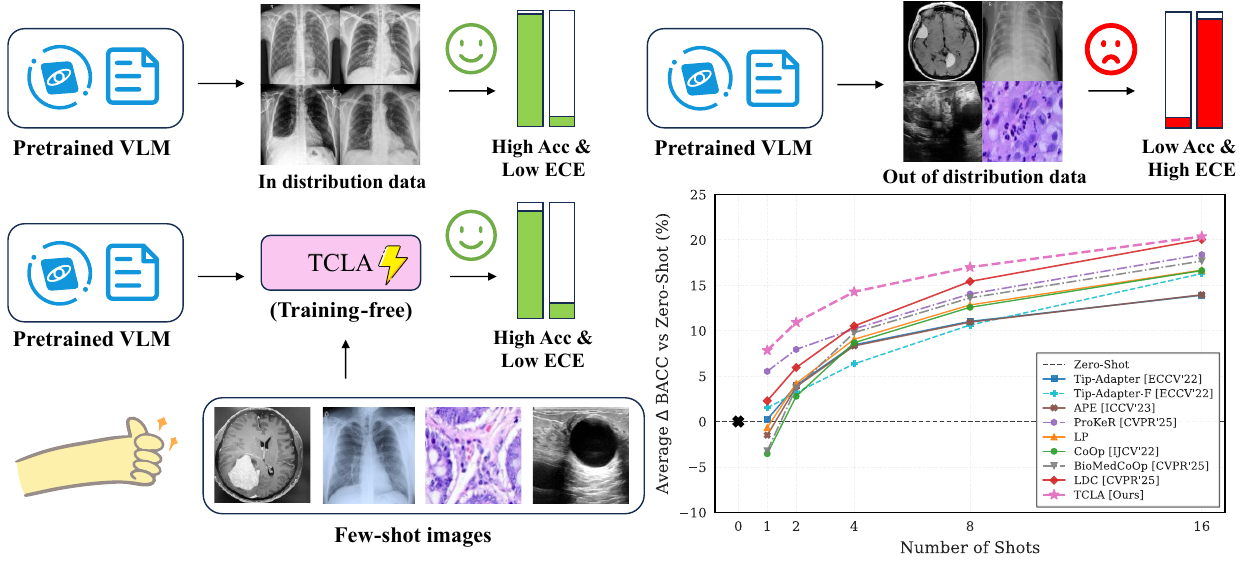}
\caption{Motivation and balanced-accuracy improvements compared with ZS, averaged over nine datasets. Results including 1-16 shots are averaged over 50 runs with different random seeds.}
\label{fig1}
\end{figure}

Motivated by these challenges, we propose TCLA (Training-free Class-wise Logit Adaptation), a training-free method that adapts the logits of medical VLMs for downstream diagnostic tasks. 
TCLA computes class-wise, layer-adaptive prototypes from few-shot support sets, constructs prototype-based correction bases, and estimates a closed-form residual mapping to adapt zero-shot logits without updating model parameters. 
We evaluate TCLA across multiple imaging modalities, including X-ray, computed tomography (CT), magnetic resonance imaging (MRI), ultrasound, and histopathology, covering a diverse range of clinical scenarios. 
Fig.~\ref{fig1} illustrates the motivation of our approach and summarizes the balanced accuracy (BACC) improvements obtained by applying TCLA to BioMedCLIP, averaged over all datasets. 
The code is available at \url{https://github.com/SkyCol/TCLA}.

\section{Related Work}
\subsection{Vision-Language Pre-trained Models}
Contrastive Language-Image Pre-training (CLIP) is a milestone for vision-language models (VLMs) that learns joint representations of images and text through large-scale contrastive pre-training \cite{Radford2021}. CLIP-based models \cite{Radford2021, Jia2021ALIGN} typically consist of an image encoder and a text encoder that separately extract features from images and text. A projection layer is usually applied to the image features to ensure that their dimensionality matches that of the text features, enabling a shared embedding space. Such VLMs have shown powerful Zero-Shot Learning (ZSL) and Few-Shot Learning (FSL) capabilities across downstream tasks.
 
Large-scale, image-caption paired datasets in medical domain have enabled Vision-Language Pretraining (VLP). PubMedCLIP \cite{Eslami2023} used one million image-caption pairs from PMC-OA to train an image-captioning model with a dual contrastive learning and masked language modeling. MedCLIP \cite{Wu2023} trained CLIP with generated pairs combinatorially from existing medical datasets. BioMedCLIP \cite{Zhang2023BioMedCLIP} introduced PMC-15M dataset which contains 15 million biomedical image-text pairs collected from 4.4 million scientific articles, and trained BiomedCLIP as a multi-modal foundation model. More recently,  BIOMEDICA \cite{Lozano2025} further released a more comprehensive dataset, containing 24 million image–caption pairs from 6 million scientific papers for large-scale biomedical VLP. 

\subsection{Few-shot Adaptation for VLMs}
Real-world data is usually out-of-distribution (OOD), i.e., unseen disease categories or differ in imaging distribution due to variations in hospitals, scanners, or patient populations, which limits the scalability of these models to new data. To improve model robustness and performance under such OOD conditions, few-shot adaptation methods have been proposed to adapt VLMs while only using a few labeled samples. 
\subsubsection{Training-based Few-shot Adaptation}
Training-based few-shot learning methods for CLIP typically freeze the entire image backbone and text backbone, and add adaptive layers to achieve Prompt Learning (PL), Feature Adaptation (FA) or Logit Adaptation (LA). PL methods adapt prompts by inserting a set of learnable tokens in the original text prompt at the input of CLIP text encoder. Methods such as CoOp \cite{Zhou2022} and KgCoOp \cite{Yao2023}, optimize these tokens using cross-entropy loss, shifting the text embedding in the semantic space to better match the few-shot image distribution. FA methods, including LoRA \cite{Hu2022} and CLIP Adapter \cite{Gao2024}, adapt features extracted from the image encoder, text encoder, or both, by adding learnable layers inside or after the CLIP backbone. The residual design of adapters in FA methods plays a critical role in few-shot settings, enabling adaptation without overwriting CLIP’s pretrained representations. LA methods directly adjust the CLIP zero-shot logits: Tip-Adapter constructs a support-set cache from few-shot image features and combines cache-based logits with zero-shot CLIP logits in a training-free manner, while Tip-Adapter-F \cite{Zhang2022TipAdapter-F} fine-tunes the cache keys through supervised training. LDC \cite{Li2025} adapts logits using image features from multiple layers of the visual encoder, to adjust logits to achieve inter-class deconfusion.  

A recent paper compares several few-shot adaptation methods of Medical VLMs indicates the most previous method Linear Probe (LP) can outperform recent methods in certain cases \cite{Kravets2025}. We found one main reason is adaptation model parameters, including residual weight parameter and learnable layer parameter, can fail when trained with extremely few samples. In this case, training-free methods have the adavantages of trustworthy improvement, faster, and easy to deploy.

\subsubsection{Training-free Few-shot Adaptation}
Training-free few-shot adaptation methods have recently attracted increasing attention as lightweight alternatives to parameter-efficient fine-tuning. These methods keep the pretrained vision-language model fixed and use a small labeled support set to adjust predictions at inference time. Tip-Adapter \cite{Zhang2022} is a representative cache-based approach, which stores few-shot visual features as a cache model and combines cache-based logits with the original CLIP zero-shot logits. APE \cite{Zhu2023APE} further improves adaptation by selecting and aggregating informative support examples, reducing the influence of noisy or less representative samples. More recently, ProKeR \cite{Bendou2025Proker} introduced a kernel perspective on few-shot adaptation and formulates support-set matching as a kernel-based prediction problem to correct logits.

\section{Methods}
\begin{figure}[t]
\centering 
\includegraphics[width=\textwidth, 
                 height=\textheight, 
                 keepaspectratio,      
                 trim=0cm 18.8cm 0cm 0.1cm, % 左 下 右 上
                 clip]{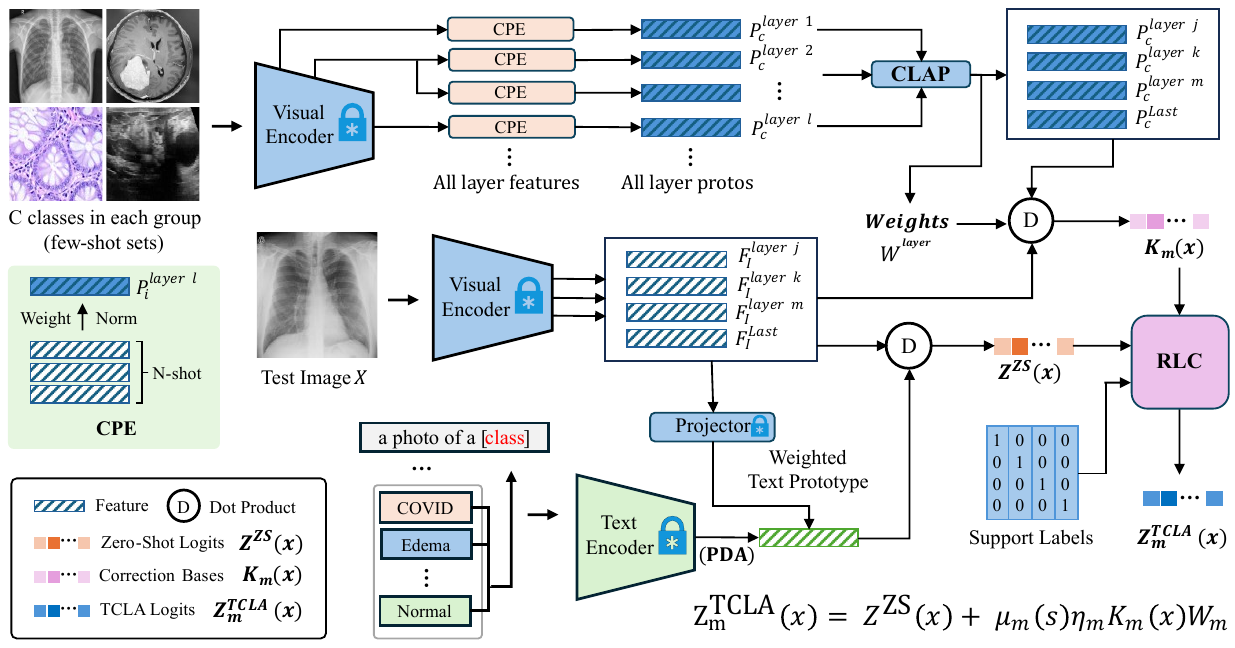}
\caption{Overall architecture of TCLA. CPE extracts class-wise prototypes, CLAP builds class-wise multi-layer prototype correction bases from the few-shot support set, , PDA constructs adapted text prototypes for zero-shot prediction. RLC estimates a closed-form residual mapping from support-set prototype responses to zero-shot logit residuals, producing final TCLA logits through prototype-guided residual correction.
} 
\label{fig2}\end{figure}
To adapt and boost the performance of Medical VLMs on downstream diagnosis tasks, we present the Training-free Class-wise Logit Adaptation (TCLA) framework. An overview of the method is illustrated in Fig.~\ref{fig2}.

\subsection{Class-wise Layer-Adaptive Prototype (CLAP).}

CLAP constructs class-wise visual prototypes from frozen image features and
selects discriminative visual layers for the target few-shot task. Given a
support set, let $f_i^{(l)}$ denote the normalized visual feature of the
$i$-th support image at layer $l$. For each class $c$ and layer $l$, we compute
the prototype
\begin{equation}
p_c^{(l)}
=
\operatorname{Proto}
\left(
\{f_i^{(l)}: y_i=c\}
\right),
\end{equation}
where $\operatorname{Proto}(\cdot)$ denotes a Mahalanobis-weighted prototype
estimator. For each support feature $f_i^{(l)}$, we compute its
Mahalanobis distance to the class center as
$d_i^{(l)}=\sqrt{(f_i^{(l)}-\bar f_c^{(l)})^\top
(\Sigma_c^{(l)})^{-1}(f_i^{(l)}-\bar f_c^{(l)})}$ and assign the sample weight
$\pi_i^{(l)}=\exp(-d_i^{(l)})/\sum_{j:y_j=c}\exp(-d_j^{(l)})$. The prototype is
then obtained by weighted aggregation,
$p_c^{(l)}=\sum_{i:y_i=c}\pi_i^{(l)}f_i^{(l)}$. In the one-shot case, the
single normalized support feature is directly used as the class prototype.

To select informative layers, we measure the inter-class separability of
prototypes at each layer. Let
$P^{(l)}=[p_1^{(l)},\dots,p_C^{(l)}]$ denote the class prototype matrix. We
compute the average inter-class cosine similarity
\begin{equation}
\bar{s}^{(l)}
=
\frac{2}{C(C-1)}
\sum_{i<j}
\cos
\left(
p_i^{(l)},p_j^{(l)}
\right),
\end{equation}
and define the layer score as
\begin{equation}
\operatorname{score}(l)=1-\bar{s}^{(l)}.
\end{equation}
The top-$K$ layers with the highest scores are selected as
$\mathcal{L}^{*}$, and their aggregation weights are normalized by
\begin{equation}
\omega^{(l)}
=
\frac{
\operatorname{score}(l)
}{
\sum_{m\in\mathcal{L}^{*}}\operatorname{score}(m)
},
\qquad
l\in\mathcal{L}^{*}.
\end{equation}

For a test image $x$, CLAP computes its affinity to class prototypes at each
selected layer:
\begin{equation}
A^{(l)}(x)
=
f^{(l)}(x)(P^{(l)})^\top .
\end{equation}
The affinity is converted into a sharp prototype correction basis:
\begin{equation}
K_{\mathrm{sharp}}^{(l)}(x)
=
\exp
\left[
\beta
\left(
A^{(l)}(x)-1
\right)
\right],
\end{equation}
where $\beta$ controls the sharpness of prototype matching. The multi-layer
sharp prototype basis is then
\begin{equation}
K_{\mathrm{sharp}}(x)
=
\sum_{l\in\mathcal{L}^{*}}
\omega^{(l)}
K_{\mathrm{sharp}}^{(l)}(x).
\end{equation}

This basis provides a task-specific correction basis for the residual logit correction
stage, allowing support-set evidence to adapt zero-shot logits.

\subsection{Image-guided Prompt Distribution Alignment (PDA).}

PDA refines text prototypes by reweighting prompt embeddings according
to their compatibility with projected support image features. For class $c$, let
$\{t_{c,j}\}_{j=1}^{P}$ denote the normalized embeddings of $P$ prompt
templates. For each selected layer $l\in\mathcal{L}^{*}$, let
$\{\tilde{x}_{c,i}^{(l)}\}_{i=1}^{N_c}$ denote the projected and normalized
support image features in the text-aligned embedding space. We first compute
the layer-wise visual prototype
\begin{equation}
\tilde{p}_{c}^{(l)}
=
\frac{
\sum_{i=1}^{N_c}\tilde{x}_{c,i}^{(l)}
}{
\left\|
\sum_{i=1}^{N_c}\tilde{x}_{c,i}^{(l)}
\right\|_2
}.
\end{equation}
The compatibility between prompt $t_{c,j}$ and the visual prototype is
$s_{c,j}^{(l)}=\langle \tilde{p}_{c}^{(l)},t_{c,j}\rangle$. We then compute
the prompt weight as
\begin{equation}
w_{c,j}^{(l)}
=
\frac{
\exp(\beta_t s_{c,j}^{(l)})
}{
\sum_{k=1}^{P}\exp(\beta_t s_{c,k}^{(l)})
},
\end{equation}
where $\beta_t$ controls the sharpness of prompt weighting. The layer-wise text
prototype is obtained by
\begin{equation}
t_c^{(l)}
=
\sum_{j=1}^{P}
w_{c,j}^{(l)} t_{c,j}.
\end{equation}
Finally, PDA aggregates the text prototypes across selected layers using the
CLAP layer weights:
\begin{equation}
t_c
=
\sum_{l\in\mathcal{L}^{*}}
\omega^{(l)} t_c^{(l)}.
\end{equation}

The resulting text prototype $t_c$ is used as the class text representation
for computing zero-shot logits.

\subsection{Prototype-guided Residual Logit Correction (RLC).}

Thanks to the kernel perspective of ProKeR~\cite{Bendou2025Proker}, RLC corrects the PDA zero-shot logits using support-set prototype evidence. It learns a closed-form residual
mapping from CLAP responses to logit corrections. Based on the CLAP affinity $A^{(l)}(x)$, RLC constructs two layer-wise
prototype kernels:
\begin{equation}
K_{\mathrm{sharp}}^{(l)}(x)
=
\exp
\left[
\beta
\left(
A^{(l)}(x)-1
\right)
\right],
\qquad
K_{\mathrm{soft}}^{(l)}(x)
=
\frac{
A^{(l)}(x)+1
}{2}.
\end{equation}
The sharp kernel emphasizes confident prototype matches, while the soft kernel
provides a smoother prototype response.

TCLA uses the same residual correction formulation with two adaptation modes.
The aggressive mode directly uses the sharp prototype kernel, while the
conservative mode interpolates between the soft and sharp kernels according to
prototype reliability to better preserve the zero-shot prior:
\begin{equation}
\begin{aligned}
K_{\mathrm{agg}}(x)
&=
\sum_{l\in\mathcal{L}^{*}}
\omega^{(l)}
K_{\mathrm{sharp}}^{(l)}(x),\\
K_{\mathrm{con}}(x)
&=
\sum_{l\in\mathcal{L}^{*}}
\omega^{(l)}
\left[
(1-r_c^{(l)})K_{\mathrm{soft}}^{(l)}(x)
+
r_c^{(l)}K_{\mathrm{sharp}}^{(l)}(x)
\right].
\end{aligned}
\end{equation}
The reliability of class prototype $p_c^{(l)}$ is computed from the compactness
of support features around the prototype:
\begin{equation}
v_c^{(l)}
=
\frac{1}{K_c}
\sum_{i:y_i=c}
\left\|
f_i^{(l)}-p_c^{(l)}
\right\|_2^2,
\qquad
r_c^{(l)}
=
\frac{K_c-1}{K_c}
\exp
\left(
-\gamma v_c^{(l)}
\right),
\end{equation}
where $K_c$ is the number of support samples for class $c$. Thus, one-shot
prototypes are treated conservatively, while more
support samples are allowed to contribute sharper corrections.

For either mode $m\in\{\mathrm{agg},\mathrm{con}\}$, let
$K_m(x)\in\mathbb{R}^{C}$ denote the computed prototype correction basis.
For the support set $\mathcal{S}=\{(x_i,y_i)\}_{i=1}^{N}$, we collect the support prototype responses as
\begin{equation}
K_{\mathcal{S}}^{m}
=
\begin{bmatrix}
K_m(x_1)\\
\cdots\\
K_m(x_N)
\end{bmatrix}
\in\mathbb{R}^{N\times C}.
\end{equation}
Let $Z_{\mathcal{S}}^{\mathrm{ZS}}\in\mathbb{R}^{N\times C}$ denote the
zero-shot logits of all support images. Given the one-hot label matrix
$Y_{\mathcal{S}}\in\mathbb{R}^{N\times C}$, the support residual is
\begin{equation}
R_{\mathcal{S}}
=
Y_{\mathcal{S}}
-
Z_{\mathcal{S}}^{\mathrm{ZS}}.
\end{equation}

We estimate a closed-form mapping from the prototype correction basis to the
zero-shot residual by ridge regression:
\begin{equation}
W_m
=
\left(
(K_{\mathcal{S}}^{m})^\top K_{\mathcal{S}}^{m}
+
\lambda I
\right)^{-1}
(K_{\mathcal{S}}^{m})^\top
R_{\mathcal{S}},
\end{equation}
where $\lambda=$ 1.0 is a fixed ridge regularization coefficient. The predicted support
residual is $\hat{R}_{\mathcal{S}}^{m}=K_{\mathcal{S}}^{m}W_m$.

While $W_m$ captures the class-wise correction pattern, ridge regression may
produce a residual with a mismatched magnitude. We therefore estimate a scalar
magnitude coefficient by projecting the target residual onto the predicted
residual:
\begin{equation}
\eta_m
=
\frac{
\left\langle
R_{\mathcal{S}},
\hat{R}_{\mathcal{S}}^{m}
\right\rangle_F
}{
\left\|
\hat{R}_{\mathcal{S}}^{m}
\right\|_F^2
+
\epsilon
}.
\end{equation}
Equivalently, $\eta_m$ is the closed-form solution that minimizes
$\|R_{\mathcal{S}}-\eta_m\hat{R}_{\mathcal{S}}^{m}\|_F^2$. Thus, $W_m$
models the correction pattern, while $\eta_m$ calibrates the overall residual
strength.

For an inference image $x$, the final TCLA logits are computed as
\begin{equation}
\boxed{
Z_m^{\mathrm{TCLA}}(x)
=
Z^{\mathrm{ZS}}(x)
+
\mu_m(s)
\eta_m
K_m(x)W_m
}
\end{equation}
where $s$ denotes the number of shots per class. Here, $W_m$ and $\eta_m$ are
estimated from the support set.

The coefficient $\mu_m(s)$ controls the mode-level correction strength:
\begin{equation}
\mu_{\mathrm{agg}}(s)=1,
\qquad
\mu_{\mathrm{con}}(s)=\frac{s}{s+1}.
\end{equation}
We use the conservative mode for CT
and MRI datasets, as few-shot prototypes in
these modalities may capture slice- or sequence-level
variations. The mode is predefined by modality and does not use a
validation set.

% This residual formulation keeps all encoders frozen and avoids validation-set
% tuning. The aggressive mode is suitable when support prototypes provide strong
% task-specific evidence, whereas the conservative mode is more appropriate when
% few-shot prototypes may capture acquisition- or slice-level variation.

\section{Experiment}
\subsection{Experimental Setup}
\subsubsection{Datasets} We evaluate model performance on nine datasets in Radiology, including various modalities: X-ray — Chest X-Ray Pneumonia~\cite{Kermany2018} (CXRP), COVIDx~\cite{Wang2020} (COVID), Tuberculosis Shenzhen~\cite{Jaeger2014} (TBSZ), CheXpert~\cite{Irvin2019,Shakeri2024} (CXP\textsubscript{5×1000}). CT — CTKidney~\cite{Islam2022} (CTKi), SARS COV2 CtScan~\cite{Soares2020} (SCCT). MRI — Brain Tumor MRI~\cite{Nickparvar2021} (BTMRI). Ultrasound — BUSI~\cite{AlDhabyani2020} (BUSI).
Histopathology — LC25000~\cite{Borkowski2019} (LC25k). These datasets also cover a wide variety of diseases, such as Pneumonia, COVID-19, and Cancer.

\smallskip
\subsubsection{Baselines.} We adopt BioMedCLIP \cite{Zhang2023BioMedCLIP} as the fundamental model for few-shot adaptation. BioMedCLIP follows the original CLIP framework and is pretrained on PMC-15M, which contains 15 million biomedical image–text pairs collected from publicly available full-text articles. We compare the proposed TCLA with both training-free (Tip-Adapter \cite{Zhang2022}, APE \cite{Zhu2023APE}, ProKeR \cite{Bendou2025Proker}) and training-based methods (LP \cite{Yosinski2014}, Tip-Adapter-F \cite{Zhang2022TipAdapter-F} CoOp \cite{Zhou2022}, BiomedCoOp \cite{Koleilat2025BiomedCoOp} and LDC \cite{Li2025}).

\subsubsection{Evaluation Metrics}
We report balanced accuracy (BACC) as the primary metric. 
BACC computes the accuracy within each ground-truth class and then averages these class-wise values, which prevents majority classes from dominating the evaluation. 
This makes it more suitable than overall accuracy for class-imbalanced medical image datasets. 
To reduce the influence of sampling variance caused by limited few-shot examples, we report the mean and standard deviation over multiple random support set splits.

\subsection{Comparative Evaluation}
\begin{figure}[!ht]
\centering 
\includegraphics[width=1\textwidth, 
                 height=1\textheight,
                 keepaspectratio,     
                 trim=0cm 0cm 0cm 0cm, 
                 clip]{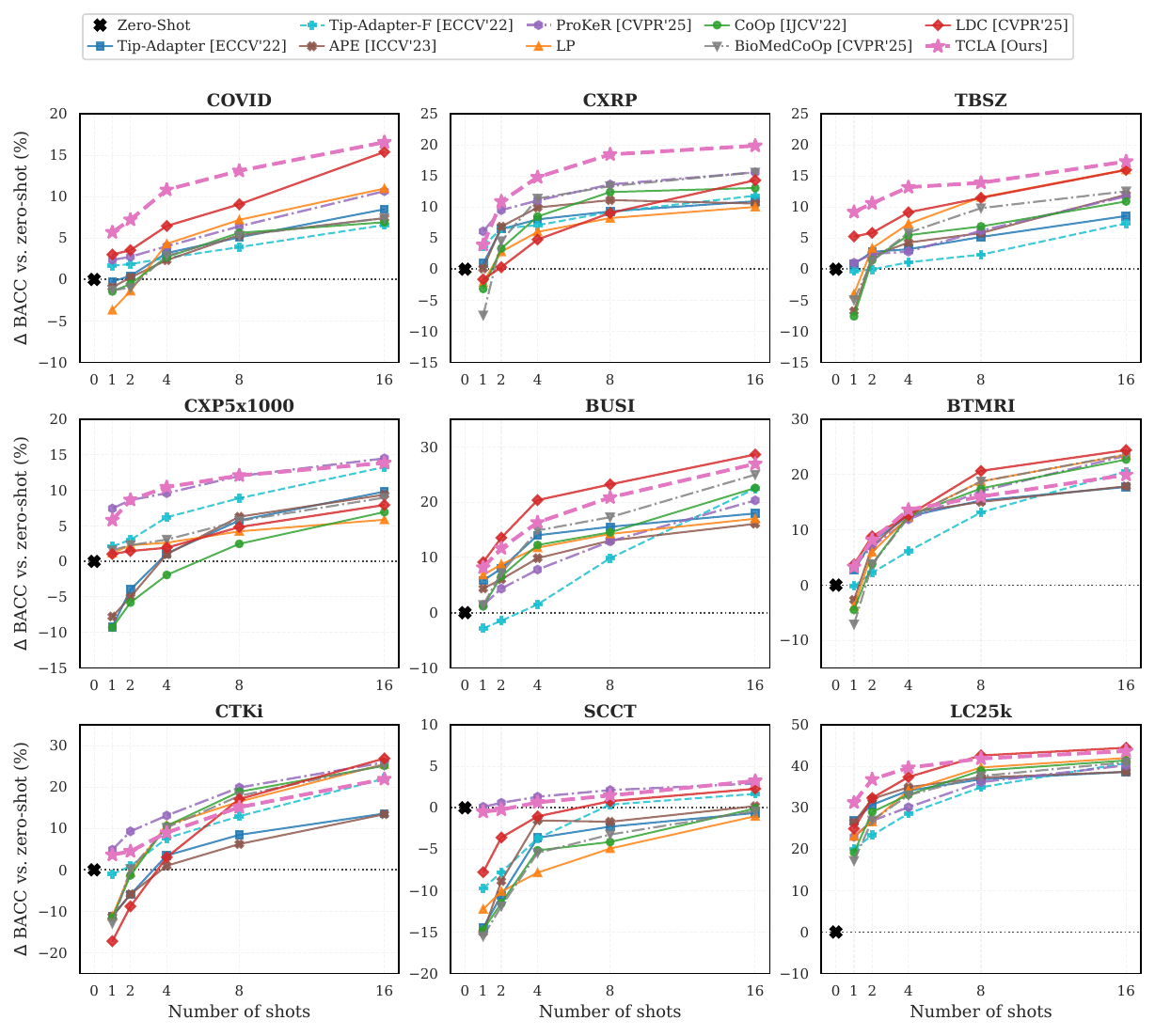}
\caption{Few-shot improvement over zero-shot balanced accuracy across nine medical imaging datasets.} 
% \caption{Overall model performance} 
\label{fig3}
\end{figure}
\paragraph{Performance overview}

Fig.~\ref{fig3} shows the few-shot adaptation performance
of TCLA with both training-free and training-based baselines of nine datasets. TCLA consistently improves with more support samples and achieves highly competitive performance on most datasets. The gains are particularly evident on COVID, CXRP, TBSZ, CXP$_{5\times1000}$, BUSI, and LC25k, where TCLA maintains a clear advantage over both training-free and training-based baselines. 
% TCLA shows consistent gains in the low-shot regime,
% indicating that prototype-guided residual correction provides a strong
% training-free alternative to optimization-based adaptation.

\subsubsection{Comparison with training-free methods}

\begin{table}[!ht]
\centering
\caption{Balanced accuracy compared with training-free methods.}
\label{tab:training_free_bacc}
\setlength{\aboverulesep}{0pt}
\setlength{\belowrulesep}{0pt}
\footnotesize
\renewcommand{\arraystretch}{0.8}
\setlength{\tabcolsep}{2.5pt}
\vspace{-0.8em}
\begin{tabular}{c|c|p{0.06\textwidth}p{0.06\textwidth}p{0.06\textwidth}p{0.06\textwidth}p{0.06\textwidth}p{0.06\textwidth}p{0.06\textwidth}p{0.06\textwidth}p{0.06\textwidth}}
\toprule
\rotatebox{45}{\textbf{Shot}} & \rotatebox{45}{\textbf{Method}} 
& \rotatebox{45}{\textbf{COVID}} 
& \rotatebox{45}{\shortstack{\textbf{CXRP}}} 
& \rotatebox{45}{\shortstack{\textbf{TBSZ}}} 
& \rotatebox{45}{\shortstack{\textbf{CXP}\\\textsubscript{5$\times$1000}}}
& \rotatebox{45}{\textbf{BUSI}} 
& \rotatebox{45}{\textbf{BTMRI}} 
& \rotatebox{45}{\textbf{CTKi}} 
& \rotatebox{45}{\textbf{SCCT}}
& \rotatebox{45}{\textbf{LC25k}} \\
\midrule

\multicolumn{2}{c|}{\textbf{Zero-shot}} 
& 55.64 & 73.62 & 64.38 & 41.64 & 44.46 & 59.19 & 54.15 & 80.21 & 49.49 \\
\midrule

& Tip-Adapter & 55.39 & 74.55 & 65.22 & 32.43 & 50.26 & 61.88 & 43.01 & 65.73 & 76.36\\
\text{1} 
& APE & 54.69 & 73.73 & 57.62 & 33.87 & 48.84 & 56.51 & 42.89 & 65.18 & 75.96\\ 
& ProKeR & 58.00 & \textbf{79.71} & 65.40 & \textbf{49.09} & 45.86 & \textbf{62.65} & \textbf{59.02} & \textbf{80.32} & 72.48\\
& TCLA(ours) & \textbf{61.34} & 77.52 & \textbf{73.57} & 47.45 & \textbf{52.63} & 62.50 & 57.79 & 79.73 & \textbf{80.77}\\
\midrule

& Tip-Adapter & 56.08 & 80.07 & 67.16 & 37.67 & 52.47 & 67.15 & 48.21 & 69.62 & 80.21\\
\text{2} 
& APE & 55.89 & 80.47 & 66.50 & 36.62 & 50.50 & 66.17 & 48.30 & 71.36 & 81.65\\
& ProKeR & 58.39 & 83.12 & 66.84 & 50.27 & 48.81 & 66.17 & \textbf{63.46} & \textbf{80.81} & 76.36\\
& TCLA(ours) & \textbf{62.86} & \textbf{84.51} & \textbf{74.97} & \textbf{50.30} & \textbf{56.12} & \textbf{67.36} & 58.66 & 79.98 & \textbf{86.34}\\
\midrule

& Tip-Adapter & 58.84 & 81.63 & 67.59 & 42.75 & 58.48 & 71.65 & 57.64 & 76.59 & 83.64\\
\text{4} 
& APE & 57.97 & 83.54 & 68.66 & 42.65 & 54.33 & 72.29 & 55.16 & 78.68 & 84.45\\
& ProKeR & 59.65 & 84.67 & 67.20 & 51.25 & 52.26 & 71.09 & \textbf{67.28} & \textbf{81.54} & 79.62\\
& TCLA(ours) & \textbf{66.47} & \textbf{88.41} & \textbf{77.57} & \textbf{52.10} & \textbf{60.80} & \textbf{72.83} & 63.15 & 80.86 & \textbf{89.16}\\
\midrule

& Tip-Adapter & 60.73 & 82.91 & 69.56 & 47.41 & 60.01 & 74.50 & 62.61 & 77.93 & 86.36\\
\text{8} 
& APE & 61.00 & 84.73 & 70.19 & 47.93 & 57.51 & 74.24 & 60.39 & 78.52 & 86.78\\
& ProKeR & 62.06 & 87.25 & 70.56 & 53.66 & 57.38 & \textbf{76.20} & \textbf{74.04} & \textbf{82.34} & 85.68\\
& TCLA(ours) & \textbf{68.76} & \textbf{92.08} & \textbf{78.28} & \textbf{53.74} & \textbf{65.32} & 75.21 & 69.28 & 81.69 & \textbf{91.36}\\
\midrule

& Tip-Adapter & 64.12 & 84.55 & 72.93 & 51.47 & 62.43 & 76.95 & 67.76 & 79.58 & 88.16\\
\text{16} 
& APE & 63.05 & 84.18 & 76.36 & 51.03 & 60.54 & 77.08 & 67.59 & 80.41 & 88.16\\
& ProKeR & 66.30 & 89.20 & 76.11 & \textbf{56.14} & 64.79 & \textbf{82.62} & \textbf{80.09} & 83.14 & 89.74\\
& TCLA(ours) & \textbf{72.21} & \textbf{93.48} & \textbf{81.70} & 55.51 & \textbf{71.41} & 79.14 & 76.09 & \textbf{83.46} & \textbf{93.17}\\
\bottomrule
\end{tabular}
\end{table}

Table~\ref{tab:training_free_bacc} compares TCLA with representative training-free adaptation methods, as all reported results are averaged over 50 independent runs for statistical reliability. TCLA achieves the best performance on most datasets and shot settings. Compared with Tip-Adapter and APE, TCLA shows more stable improvements across heterogeneous medical datasets, indicating that multi-layer class-wise prototype-guided residual logit correction is more effective than relying only on cache-based retrieval or prompt refinement. ProKeR is a strong baseline, especially on datasets such as CTKi and SCCT, where it remains competitive or slightly better in several settings. Nevertheless, TCLA achieves stronger average performance across shots and shows particularly large improvements on COVID, TBSZ, BUSI, and LC25k.

The advantage of TCLA becomes more stable as the number of support examples increases. For example, on COVID, TCLA improves from 61.34 in the 1-shot setting to 72.21 in the 16-shot setting, and on BUSI it increases from 52.63 to 71.41. Similar trends are observed on CXRP and LC25k, where TCLA reaches 93.48 and 93.17 in the 16-shot setting, respectively. These results indicate that TCLA can effectively exploit additional few-shot support samples while still maintaining strong performance in the extremely few-shot regime.

\subsubsection{Comparison with training-based methods}

\begin{table}[!ht]
\centering
\caption{Balanced accuracy compared with training-based methods.}
\label{tab:training_based_bacc}
\setlength{\aboverulesep}{0pt}
\setlength{\belowrulesep}{0pt}
\footnotesize
\renewcommand{\arraystretch}{0.8}
\setlength{\tabcolsep}{2.5pt}
\vspace{-0.8em}
\begin{tabular}{c|c|p{0.06\textwidth}p{0.06\textwidth}p{0.06\textwidth}p{0.06\textwidth}p{0.06\textwidth}p{0.06\textwidth}p{0.06\textwidth}p{0.06\textwidth}p{0.06\textwidth}}
\toprule
\rotatebox{45}{\textbf{Shot}} & \rotatebox{45}{\textbf{Method}} 
& \rotatebox{45}{\textbf{COVID}} 
& \rotatebox{45}{\shortstack{\textbf{CXRP}}} 
& \rotatebox{45}{\shortstack{\textbf{TBSZ}}} 
& \rotatebox{45}{\shortstack{\textbf{CXP}\\\textsubscript{5$\times$1000}}}
& \rotatebox{45}{\textbf{BUSI}} 
& \rotatebox{45}{\textbf{BTMRI}} 
& \rotatebox{45}{\textbf{CTKi}} 
& \rotatebox{45}{\textbf{SCCT}}
& \rotatebox{45}{\textbf{LC25k}} \\
\midrule

\multicolumn{2}{c|}{\textbf{Zero-shot}} 
& 55.64 & 73.62 & 64.38 & 41.64 & 44.46 & 59.19 & 54.15 & 80.21 & 49.49 \\
\midrule

& LP & 51.94 & 71.49 & 60.39 & 42.84 & 51.27 & 55.35 & 42.98 & 67.97 & 72.60\\
& Tip-Adapter-F & 57.30 & 77.32 & 64.07 & 43.77 & 41.64 & 59.05 & 53.17 & 70.49 & 69.55\\
\text{1}
& CoOp & 54.20 & 70.45 & 56.79 & 32.33 & 45.62 & 54.70 & 42.63 & 65.49 & 68.70\\
& BiomedCoOp & 54.37 & 66.19 & 59.41 & 43.21 & 45.86 & 52.07 & 41.24 & 64.70 & 66.68\\
& LDC & 58.65 & 71.98 & 69.64 & 42.67 & \textbf{53.61} & \textbf{62.93} & 36.95 & 72.43 & 74.44\\
& TCLA(ours) & \textbf{61.34} & \textbf{77.52} & \textbf{73.57} & \textbf{47.45} & 52.63 & 62.50 & \textbf{57.79} & \textbf{79.73} & \textbf{80.77}\\
\midrule

& LP & 54.26 & 76.43 & 67.75 & 43.91 & 53.23 & 65.16 & 53.55 & 70.10 & 76.10\\
& Tip-Adapter-F & 57.50 & 80.16 & 64.33 & 44.68 & 42.99 & 61.43 & 55.03 & 72.42 & 72.87\\
\text{2}
& CoOp & 55.11 & 76.95 & 65.84 & 35.85 & 51.06 & 63.01 & 52.84 & 68.70 & 78.46\\
& BiomedCoOp & 54.62 & 78.14 & 65.70 & 43.92 & 51.67 & 62.96 & 54.35 & 68.31 & 76.47\\
& LDC & 59.18 & 73.94 & 70.22 & 43.10 & \textbf{58.02} & \textbf{68.01} & 45.35 & 76.62 & 81.85\\
& TCLA(ours) & \textbf{62.86} & \textbf{84.51} & \textbf{74.97} & \textbf{50.30} & 56.12 & 67.36 & \textbf{58.66} & \textbf{79.98} & \textbf{86.34}\\
\midrule

& LP & 59.95 & 79.62 & 71.68 & 44.27 & 56.23 & 71.53 & \textbf{64.95} & 72.38 & 83.67\\
& Tip-Adapter-F & 58.24 & 80.69 & 65.49 & 47.85 & 45.96 & 65.32 & 61.80 & 76.52 & 78.10\\
& CoOp & 58.35 & 82.09 & 69.85 & 39.73 & 56.71 & 71.71 & 64.77 & 75.07 & 82.63\\
\text{4}
& BiomedCoOp & 58.51 & 84.98 & 70.25 & 44.71 & 59.35 & 71.43 & 64.36 & 74.76 & 82.53\\
& LDC & 62.10 & 78.39 & 73.52 & 43.56 & \textbf{64.84} & 71.81 & 57.20 & 79.17 & 86.91\\
& TCLA(ours) & \textbf{66.47} & \textbf{88.41} & \textbf{77.57} & \textbf{52.10} & 60.80 & \textbf{72.83} & 63.15 & \textbf{80.86} & \textbf{89.16}\\
\midrule

& LP & 62.84 & 81.84 & 75.94 & 45.87 & 58.72 & 77.95 & 70.64 & 75.29 & 89.22\\
& Tip-Adapter-F & 59.55 & 82.81 & 66.71 & 50.54 & 54.32 & 72.35 & 67.11 & 80.59 & 84.43\\
\text{8}
& CoOp & 61.28 & 86.03 & 71.25 & 44.11 & 59.01 & 76.69 & \textbf{73.02} & 76.08 & 88.47\\
& BiomedCoOp & 60.94 & 86.98 & 74.20 & 47.34 & 61.75 & 77.94 & 72.02 & 76.98 & 87.15\\
& LDC & 64.72 & 82.66 & 75.85 & 46.47 & \textbf{67.68} & \textbf{79.86} & 71.24 & 81.01 & \textbf{92.11}\\
& TCLA(ours) & \textbf{68.76} & \textbf{92.08} & \textbf{78.28} & \textbf{53.74} & 65.32 & 75.21 & 69.28 & \textbf{81.69} & 91.36\\
\midrule

& LP & 66.62 & 83.63 & 80.39 & 47.51 & 61.47 & 82.78 & 79.69 & 79.19 & 91.51\\
& Tip-Adapter-F & 62.21 & 85.44 & 71.75 & 54.92 & 66.95 & 79.75 & 76.18 & 81.89 & 90.28\\
\text{16}
& CoOp & 62.55 & 86.68 & 75.28 & 48.59 & 67.04 & 81.92 & 79.29 & 80.13 & 90.91\\
& BiomedCoOp & 63.05 & 89.21 & 76.91 & 50.62 & 69.44 & 82.84 & 79.49 & 79.81 & 90.47\\
& LDC & 71.04 & 87.94 & 80.34 & 49.58 & \textbf{73.10} & \textbf{83.62} & \textbf{81.06} & 82.50 & \textbf{93.98}\\
& TCLA(ours) & \textbf{72.21} & \textbf{93.48} & \textbf{81.70} & \textbf{55.51} & 71.41 & 79.14 & 76.09 & \textbf{83.46} & 93.17\\
\bottomrule

\end{tabular}
\end{table}

Table~\ref{tab:training_based_bacc} compares TCLA with five representative training-based adaptation methods, all reported results are averaged over 50 independent runs for statistical reliability. Despite requiring no additional model parameters or gradient-based optimization, TCLA can achieve competitive or superior performance on most datasets and shot settings. Compared with CoOp and BioMedCoOp, TCLA is more stable in the fewer-shot regime, suggesting that preserving the zero-shot prior and applying residual corrections is more reliable than directly optimizing prompts from very limited number of samples. Although LDC remains competitive on some modalities such as ultrasound, MRI, and CT, TCLA provides a strong training-free alternative with robust performance across diverse medical imaging datasets.

\subsection{Extreme Low-Shot Analysis Across Modalities}
\begin{figure}[!ht]
\centering 
\includegraphics[width=1\textwidth, 
                 height=1\textheight,
                 scale=1,
                 keepaspectratio,     
                 trim=0cm 20.5cm 0cm 0cm, 
                 clip]{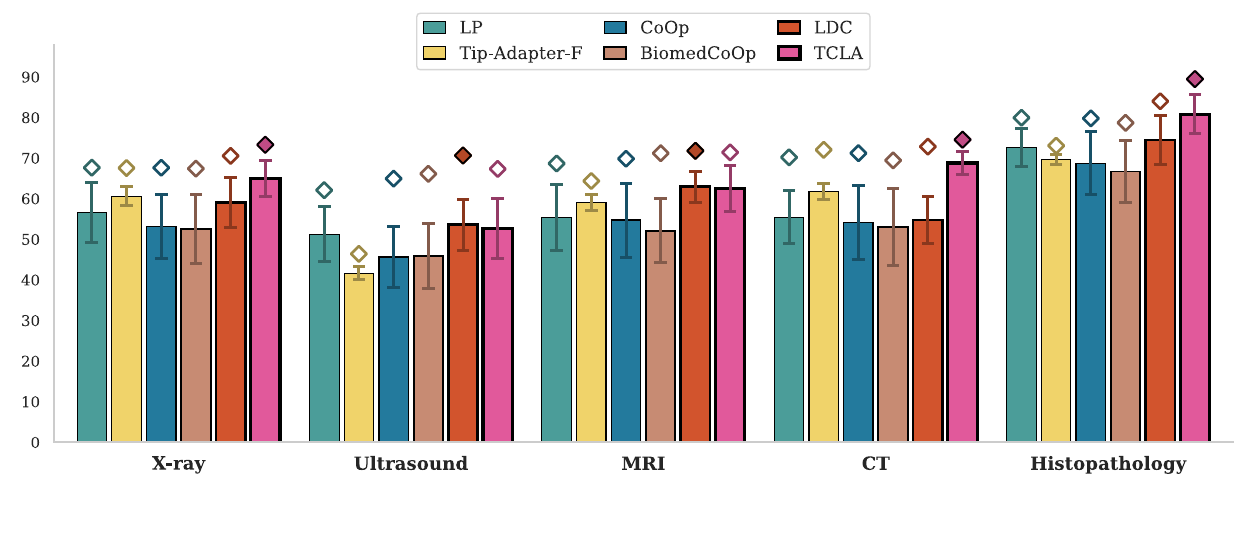}
\caption{1-shot adaptation performance across modalties. Bars indicate the mean balanced accuracy over datasets within each modality. Error bars represent the standard deviation.
Diamond markers denote the best accuracy achieved during multiple seeds.}
\label{fig4}
\end{figure}
Fig.~\ref{fig4} illustrates the performance of TCLA and training-based methods across five medical imaging modalities. 
In the extreme low-data regime, TCLA shows stable and competitive performance, indicating strong cross-modality generalization. 
Notably, TCLA outperforms the recent training-based and optimization-intensive method LDC on X-ray, CT, and histopathology datasets, which means that TCLA can be highly effective when only very limited support data are available. 
A similar trend is also observed when comparing training-based methods with the recent training-free method ProKeR in Table~\ref{tab:training_free_bacc}. 
These results suggest that training-free approaches may offer better robustness in extreme low-shot settings, where training-based methods can be sensitive to limited and variable support samples. All reported results are averaged over 50 independent runs to improve statistical reliability.

\subsection{Ablation Study}

To evaluate the contribution of each component in TCLA, we conduct an ablation study under four settings: 
(a) Base, which builds prototypes using only the last-layer features; 
(b) CLAP, which introduces class-wise layer-adaptive prototypes; 
(c) CLAP + PDA, which further applies prompt distribution alignment; and 
(d) CLAP + PDA + RLC, which corresponds to the full TCLA framework. 
The 1-shot results averaged over 10 random seeds are reported in Table~\ref{tab:modality_ablation}. 

\begin{table}[!ht]
\centering
\caption{Ablation study of TCLA components under the 1-shot setting across five medical imaging modalities.}
\label{tab:modality_ablation}
\renewcommand{\arraystretch}{0.85}
\setlength{\tabcolsep}{1.8pt}
\footnotesize
\begin{tabular}{cccccc}
\toprule
\textbf{Setting} 
& \textbf{X-ray} 
& \textbf{CT} 
& \textbf{MRI} 
& \textbf{Ultrasound} 
& \textbf{Histopathology} \\
\midrule
a
& 62.94 $\pm$ 5.16
& 62.40 $\pm$ 5.27
& 63.04 $\pm$ 3.77
& 51.28 $\pm$ 4.68
& 82.66 $\pm$ 1.86 \\

b
& 63.64 $\pm$ 5.51
& 61.14 $\pm$ 6.71
& 62.93 $\pm$ 3.55
& 52.01 $\pm$ 3.96
& 84.14 $\pm$ 1.10 \\

c
& 63.66 $\pm$ 5.44
& 61.34 $\pm$ 6.47
& 63.14 $\pm$ 3.43
& 52.21 $\pm$ 4.15
& 84.12 $\pm$ 1.07 \\

\textbf{d}
& \textbf{63.76 $\pm$ 4.30}
& \textbf{68.34 $\pm$ 3.16}
& \textbf{63.42 $\pm$ 3.61}
& \textbf{52.43 $\pm$ 3.97}
& \textbf{84.22 $\pm$ 2.31} \\
\bottomrule
\end{tabular}
\end{table}

Interestingly, the improvement from setting (a) to setting (b) is relatively small on MRI, and setting (b) even shows a slight decrease compared with (a). 
This suggests that for some slice-based data, high-level semantic features from the final layer may already capture the most relevant diagnostic information, while adding multi-layer prototypes does not necessarily bring consistent benefits. The improvements on CT from setting (c) to setting (d) is clear, which indicates the importance of conservative logit correction, which helps to preserve the original knowledge. Overall, these results show that different components contribute differently across modalities, but the full TCLA setting is suitable for most situations.

\begin{figure}[t]
\centering 
\includegraphics[width=1.02\textwidth, 
                 height=1.1\textheight,
                 scale=1.15,
                 keepaspectratio,     
                 trim=0cm 0cm 0cm 0.2cm, 
                 clip]{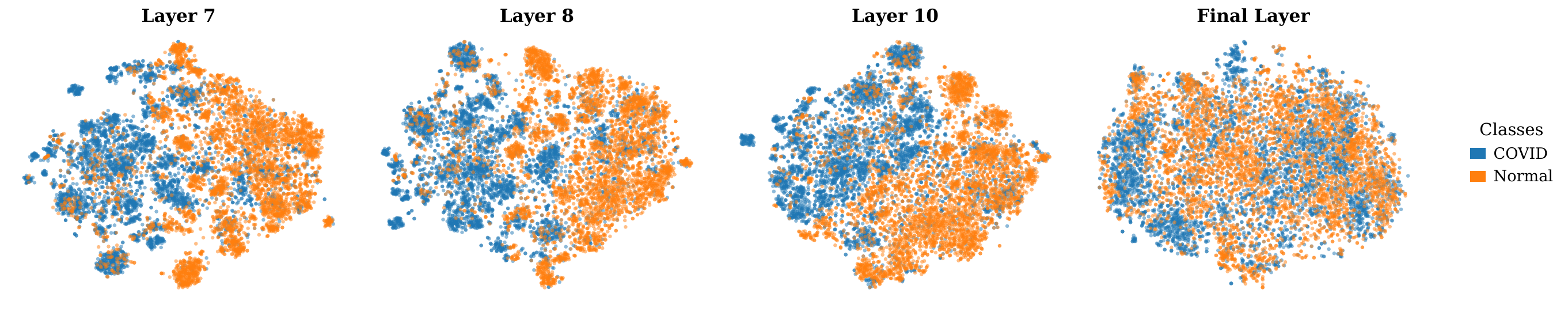}
\caption{t-SNE visualization of the image feature space for normal and COVID samples. Features from the dynamically selected intermediate layers (7, 8, and 10) exhibit improved class separability compared with the final-layer features.}
\label{fig5}
\end{figure}

Fig.~\ref{fig5} presents a two-dimensional projection of image features extracted from layers selected by CLAP. Compared to the final-layer image embedding, the feature spaces corresponding to the selected intermediate layers (layers 7, 8, and 10) exhibit noticeable improved class separability between normal and COVID samples in the COVIDx dataset.

\section{Conclusion}

In this work, we focus on the performance degradation of pretrained medical vision-language models under domain shift, where zero-shot predictions can be affected by class bias and misalignment with downstream medical datasets. To this end, we propose TCLA (Training-free Class-wise Logit Adaptation), a lightweight method for boosting medical VLMs performance when a few samples are given, without updating additional model parameters. By using these limited support samples for class-wise logit adaptation, TCLA provides a simple but robust way to correct zero-shot predictions and adapt medical VLMs to downstream diagnostic tasks.

Experiments across nine medical image datasets show that TCLA improves model performance over zero-shot inference and several existing adaptation baselines in most settings. The results suggest that even a single example can provide useful information for correcting data- or domain-specific prediction. Since TCLA does not require backpropagation or parameter updates, it can be applied to existing medical VLMs with low computational cost, making it a practical option when repeated model training is difficult or annotated data are limited.

At the same time, the current study mainly focuses on classification-level adaptation. Future work can extend toward more comprehensive adaptation of medical VLMs, especially at the vision-language alignment level. This includes better use of intermediate visual features, local region-text correspondence, retrieval, grounding, and segmentation-related transfer. Further analysis of calibration, reliability, explainability, and robustness under broader clinical domain shifts will also be important for understanding the potential and limitations of both training-free and training-based adaptation in medical vision-language modeling.

% \appendix
% \section{Example Appendix Section}


%%       \citet{<label>} ==> Lamport (1994)
%%       \citep{<label>} ==> (Lamport, 1994)
%%
% Example citation, See \citet{lamport94}.




\end{document}